\documentclass[letterpaper,10pt,conference]{ieeeconf}
\IEEEoverridecommandlockouts
\usepackage[T1]{fontenc}
\usepackage{amsmath,amssymb}
\usepackage{graphicx}
\usepackage{booktabs}
\usepackage{multirow}
\usepackage{placeins}
\usepackage{stfloats}
\usepackage{float}
\usepackage{cite}
\usepackage{xcolor}\definecolor{projectblue}{RGB}{0,174,239}
\makeatletter
\let\NAT@parse\undefined
\makeatother
\usepackage[
  colorlinks=true,
  citecolor=green,
  linkcolor=blue,
  urlcolor=black
]{hyperref}

\makeatletter
\newcommand{\compactCaptionLabel}{%
  \long\def\@makecaption##1##2{%
    \@IEEEfigurecaptionsepspace
    \parbox[t]{\hsize}{\footnotesize\noindent
      \mbox{##1.\hspace{0.45em}}##2}%
  }%
}
\makeatother

\title{\LARGE\bf Intent-Aware Whole-Body Coordination for Human--Humanoid Collaborative Transport}
\author{Zhiyuan Song, Dezhi Qin, Zhihai Bi, Jinglan Xu, Kai Chen, Hang Zhao, and Jun Ma%
\thanks{\raggedright\urlstyle{same}Zhiyuan Song, Dezhi Qin, Zhihai Bi, Jinglan Xu, Kai Chen, Hang Zhao, and Jun Ma are with the Robotics
and Autonomous Systems Thrust, The Hong Kong University of Science and
Technology (Guangzhou), Guangzhou 511453, China (e-mail:
\href{mailto:zsong142@connect.hkust-gz.edu.cn}{\nolinkurl{zsong142@connect.hkust-gz.edu.cn}};
\href{mailto:dqin642@connect.hkust-gz.edu.cn}{\nolinkurl{dqin642@connect.hkust-gz.edu.cn}};
\href{mailto:zbi217@connect.hkust-gz.edu.cn}{\nolinkurl{zbi217@connect.hkust-gz.edu.cn}};
\href{mailto:jxu367@connect.hkust-gz.edu.cn}{\nolinkurl{jxu367@connect.hkust-gz.edu.cn}};
\href{mailto:kchen916@connect.hkust-gz.edu.cn}{\nolinkurl{kchen916@connect.hkust-gz.edu.cn}};
\href{mailto:hangzhao@hkust-gz.edu.cn}{\nolinkurl{hangzhao@hkust-gz.edu.cn}}; \href{mailto:jun.ma@ust.hk}{\nolinkurl{jun.ma@ust.hk}}).}}
\hypersetup{pdftitle={Intent-Aware Whole-Body Coordination for Human–Humanoid Collaborative
Transport},pdfauthor={Zhiyuan Song, Dezhi Qin, Zhihai Bi, Jinglan Xu, Kai Chen, Hang Zhao, Jun Ma}}
\AtBeginDocument{%
  \setlength{\textfloatsep}{\baselineskip}%
  \setlength{\dbltextfloatsep}{\baselineskip}%
  \setlength{\intextsep}{\baselineskip}%
}
\begin{document}
\bstctlcite{IEEEtran:NoBibliographyURLs}
\maketitle
\thispagestyle{empty}
\pagestyle{empty}
\begin{abstract}
Human--humanoid collaborative transport combines human guidance with robotic load support to alleviate the physical demands of object transport. However, human partners convey motion intent through interaction forces, which also increase their physical burden and make reliable intent inference with reduced guiding force challenging. Furthermore, the robot must translate the inferred intent into coordinated whole-body motion to achieve object translation and tilting while maintaining balance under varying payloads. To address these limitations, we propose a hierarchical learning framework for human-humanoid collaborative transport. First, we train a privileged teacher with auxiliary intent supervision using a velocity-dependent virtual-impedance model of human guidance. The teacher is then distilled via dataset aggregation into a recurrent student that generates task-level commands from proprioceptive histories alone. Second, to execute these inferred commands, we develop an object-centric whole-body controller that incorporates object-consistent kinematic guidance and load-dependent posture guidance into policy training, enabling grasp-consistent object tilting and balanced whole-body motion across varying payloads. Experimental results demonstrate improved motion tracking and task success across varying payloads, substantially reduced human guiding force during sustained transport, and versatile collaborative transport combining planar motion and object tilting in real-world tasks such as narrow-doorway traversal and container dumping. Our project page is available at: {\hypersetup{urlcolor=projectblue}\urlstyle{same}\url{https://intent-aware-transport.github.io/}}.
\end{abstract}

\begin{keywords}
human--robot collaboration,  motion intent inference, collaborative transport, whole-body control
\end{keywords}

\section{Introduction}

\begingroup
\setlength{\parskip}{0pt}
Despite recent advances in humanoid locomotion and manipulation~\cite{gu2026humanoid,bi2026genhoi,zheng2026embracing}, fully autonomous object transport still requires reliable task-level decision-making in complex environments~\cite{wang2025hypermotion,nguyen2026hotice,bi2026alore}. Human--humanoid collaborative transport offers an alternative by combining human guidance with robotic load support, reducing the physical burden on the human partner~\cite{otani2018assistive,agravante2019carrying}. Such collaboration relies on physical interaction through a shared object. Interaction wrenches transmitted through the shared object affect the humanoid's balance and locomotion dynamics~\cite{zhang2025falcon,zhang2026iowbc}, while also conveying information about the human partner's motion intent~\cite{yang2025implicit}. Effective collaboration therefore hinges on two  challenges: reliable inference of human motion intent and coordinated whole-body motion under coupled robot--object dynamics.

\suppressfloats[t]
\begin{figure}[t]
  \vspace*{5.5pt}
  \centering
  \includegraphics[width=\columnwidth]{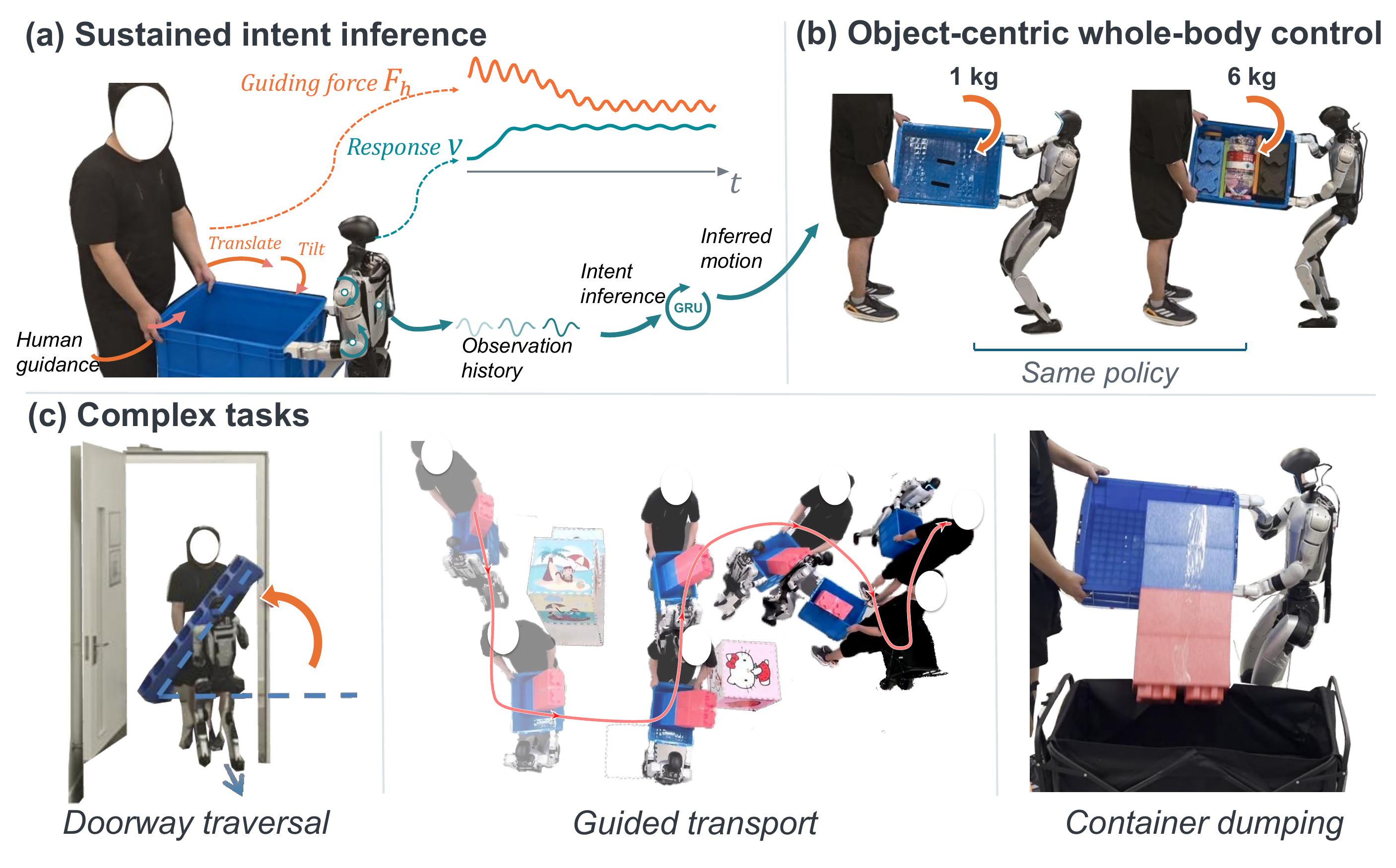}\par
  \nointerlineskip
  \compactCaptionLabel
  \caption{\textbf{The proposed framework enables a humanoid robot to infer human motion intent and perform diverse collaborative transport tasks.}}
  \label{fig:transport-overview}
\end{figure}

Previous approaches have relied on external motion-capture systems to infer human motion intent, restricting their operation to areas covered by the tracking infrastructure~\cite{sirintuna2024object}. More recently, proprioception-based methods have eliminated this dependency by estimating interaction wrenches from onboard observations and mapping them to velocity references for collaborative transport~\cite{paint}. \mbox{However, when} velocity references are directly scaled by interaction wrench magnitude, maintaining a nonzero velocity reference requires sustained human guiding force. Sustained guidance increases human effort and may impose additional interaction loads that compromise balance and reduce the actuator torque margin available for payload support. This motivates learning the human's underlying motion target from physical interaction rather than directly mapping applied forces to motion commands, enabling the robot to actively follow the intended motion with reduced sustained guiding effort.

Beyond intent inference, existing human--humanoid collaborative transport approaches primarily focus on planar translation and yaw rotation, which can be achieved through locomotion while maintaining a relatively static upper-body configuration~\cite{paint,bethala2025h2compact}. However, practical transport often extends beyond planar base motion, requiring deliberate object tilting to maneuver elongated payloads through narrow passages or tip containers for dumping. Such motions require continuous coordination between upper-body manipulation and lower-body locomotion, particularly when object tilting is combined with planar motion. Furthermore, variations in payload mass shift the center of mass (CoM) of the coupled robot--object system relative to its support polygon, with upper-body movements during object tilting further exacerbating the resulting balance challenges~\cite{dao2024box,otani2018assistive}. Task-space tracking rewards alone provide limited guidance for finding dynamically feasible whole-body motions~\cite{margolis2025softmimic}. Whole-body policy learning can guide exploration toward configurations that satisfy the shared object's bimanual kinematic constraints and maintain dynamic balance.

To address these challenges, we propose a hierarchical learning framework for human--humanoid collaborative transport that integrates whole-body execution with human motion intent inference, as illustrated in Fig.~\ref{fig:transport-overview}. At the execution layer, we develop an object-centric whole-body controller (WBC) to address the challenges of coordinating object translation and tilting while maintaining balance across varying payloads. Specifically, the WBC incorporates object-consistent kinematic guidance and load-dependent posture guidance into policy learning. The former coordinates bilateral grasp references with object tilting, while the latter accounts for CoM shifts and actuator loading across varying payloads and object orientations. At the intent inference layer, we model human guidance through velocity-dependent virtual impedance, where interaction wrenches arise from motion errors and diminish as the robot approaches the intended motion. The resulting physical interaction provides feedback for training a privileged teacher to generate task-level commands, with the human-desired motion serving as auxiliary intent supervision. We then distill the teacher through dataset aggregation (DAgger) into a recurrent student that generates task-level commands from proprioceptive histories alone.

Our main contributions are summarized as follows:
\begin{itemize}
  \item We develop an object-centric whole-body controller for versatile collaborative transport, including planar translation, yaw rotation, and deliberate object tilting. Object-consistent kinematic and load-dependent posture guidance during training promote grasp consistency and whole-body balance across varying payloads.
  \item We further introduce a human motion intent learning approach that enables sustained transport with reduced guiding effort. By combining intent supervision under virtual impedance interactions with proprioceptive policy distillation, the policy generates commands aligned with human intent, even as guiding forces diminish.
  \item Extensive experiments demonstrate the framework's effectiveness in complex collaborative transport tasks, including narrow-doorway traversal and container dumping. The results highlight improved motion coordination and task success across varying payloads, along with substantially reduced guiding effort.
\end{itemize}

\par
\endgroup

\section{Related Work}
\subsection{Whole-Body Control for Physical Interaction}
In loco-manipulation, contact forces inherently couple object dynamics with whole-body balance. Traditional whole-body optimization addresses this coupling by jointly modeling motion and force objectives within a unified framework, through constrained quadratic programming~\cite{bouyarmane2019qp,rouxel2024multicontact} or Cartesian impedance control~\cite{risiglione2022impedance}.

Moving beyond model-based optimization, reinforcement learning (RL) enables compliant contact-force regulation using only proprioceptive feedback, eliminating the need for dedicated force sensors~\cite{portela2024force}. By leveraging domain randomization and state-history observations, learned policies can robustly estimate interaction forces and adapt to varying loads during humanoid loco-manipulation~\cite{zhi2025unifp,zhang2025falcon}.

Beyond precise tracking, physical interaction requires active compliance. Recent RL approaches integrate this compliance through various mechanisms: virtual mass-spring-damper dynamics~\cite{xu2025facet, lu2025gentlehumanoid}, kinematically feasible motion augmentation~\cite{margolis2025softmimic, liu2026lac}, and force-dependent shifts in tracking goals~\cite{chen2026chip}. Recent work extends compliant whole-body control for heavier humanoids using multi-site impedance references and residual adjustments to equilibrium targets~\cite{do2026compliantwbc}. While these advances improve compliant whole-body control, their integration into intent-guided collaborative transport with deliberate object reorientation remains less explored.

\subsection{Human--Humanoid Collaborative Transport}

Physical interaction provides a natural interface for human guidance during collaborative transport~\cite{lanini2018intention}. Traditional approaches typically rely on dedicated force sensors or complex modeling, mapping wrist forces to velocity commands~\cite{bethala2025h2compact}, integrating admittance with footstep planning~\cite{agravante2019carrying}, or utilizing kinematic constraints for intent inference~\cite{ansari2021task}. Other approaches use trajectory-optimized postures~\cite{zhang2026iowbc} or vision-language-based geometric planning~\cite{zhang2026c2c}. Such methods primarily address posture optimization or geometric task planning, without directly addressing continuous motion intent inference from physical interaction. Their execution may also require online inverse kinematics (IK) or externally specified task information.

Recent learning-based approaches have enabled proprioception-driven collaborative transport with legged robots. COLA supports planar transport through residual policy distillation~\cite{du2026cola}, while PAINT estimates interaction wrenches from proprioceptive histories and employs wrench-scaled velocity references to guide policy learning~\cite{paint}. However, both approaches primarily focus on planar transport and do not address deliberate object tilting, which introduces whole-body coordination and balance challenges. Under this reference design, lower wrench magnitudes correspond to lower target velocities, even when the human intends to maintain the same motion. This motivates learning motion commands aligned with human intent while reducing reliance on wrench-scaled velocity references.
\section{Methodology}

\subsection{Overview}

\begin{figure*}[t]
  \centering
  \includegraphics[width=\textwidth]{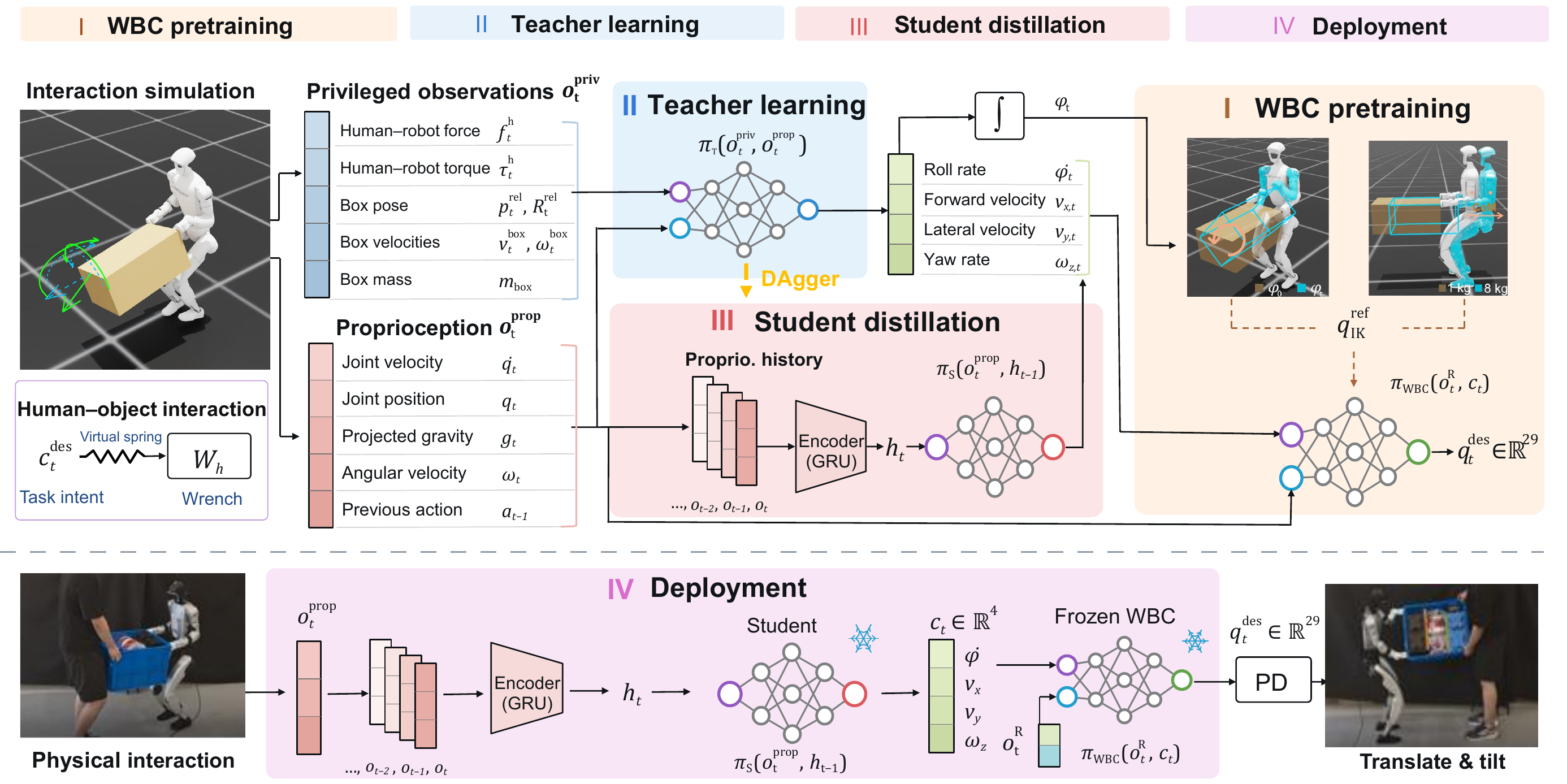}\par
  \nointerlineskip
  \caption{\textbf{Overview of the training and deployment framework.}\enspace The WBC is pretrained with kinematic and load-dependent posture guidance. A privileged teacher learns to interpret physical interaction, and a recurrent student is distilled from the teacher using proprioceptive histories. During deployment, the student and frozen WBC coordinate collaborative transport and tilting.}
  \label{fig:framework}
\end{figure*}

To enable coordinated transport and object tilting under human guidance, we adopt a hierarchical framework that decouples motion intent inference from whole-body execution, as illustrated in Fig.~\ref{fig:framework}. The framework translates the inferred human motion intent into a unified task-level motion command, which serves as the input to the whole-body controller (WBC), defined as:
\begin{equation}
\mathbf{c}_t =
[v_{x,t}^{\mathrm{cmd}},\,v_{y,t}^{\mathrm{cmd}},\,
\omega_{z,t}^{\mathrm{cmd}},\,\dot{\phi}_t^{\mathrm{cmd}}]^\top
\in\mathbb{R}^{4},
\end{equation}
where $v_{x,t}^{\mathrm{cmd}}$ and $v_{y,t}^{\mathrm{cmd}}$ denote the commanded longitudinal and lateral linear velocities of the robot base, expressed along its local $x$- and $y$-axes, respectively. The angular velocity command $\omega_{z,t}^{\mathrm{cmd}}$ specifies the desired $z$-axis component of the robot-base angular velocity expressed in the base frame. The final component $\dot{\phi}_t^{\mathrm{cmd}}$ specifies the commanded roll rate of the object relative to the robot base, enabling deliberate object tilting. Here, $\phi_t$ is the corresponding relative Euler roll angle. We denote the commands generated by the teacher and student policies as $\mathbf{c}_t^{\mathrm{T}}$ and $\mathbf{c}_t^{\mathrm{S}}$, respectively.

Training proceeds in three stages: WBC pretraining, teacher learning, and student distillation. First, the WBC is trained using kinematic references and load-dependent posture guidance to coordinate locomotion and object tilting. With the WBC frozen, a privileged teacher learns to generate motion commands from physical guidance. Finally, a recurrent student is trained through DAgger to reproduce the teacher's commands using only proprioceptive histories. During deployment, the student generates task-level commands for the frozen WBC without external force sensing. Both policies operate at 50\,Hz.

\subsection{Object-Centric Whole-Body Control}
\label{sec:wbc}

We formulate an object-centric whole-body controller that maps the four-dimensional motion command $\mathbf{c}_t$ to coordinated joint actions. Building on~\cite{zhang2025falcon}, the policy incorporates object-consistent kinematic guidance and load-dependent posture references to facilitate whole-body coordination under varying loads.

The WBC actor receives proprioceptive observations:
\begin{equation}
\mathbf{o}_t^{\mathrm{prop}} \triangleq
[\boldsymbol{\omega}_t^b,\ \mathbf{g}_t^b,\ \mathbf{q}_t-\mathbf{q}^{\mathrm{nom}},\
\dot{\mathbf{q}}_t,\ \mathbf{a}_{t-1}],
\label{eq:wbc_observation}
\end{equation}
where $\boldsymbol{\omega}_t^b$ and $\mathbf{g}_t^b$ denote the base angular velocity and projected gravity, $\mathbf{q}_t$ and $\dot{\mathbf{q}}_t$ denote the joint positions and velocities, and $\mathbf{a}_{t-1}$ denotes the previous action. The actor maintains an observation history (\(l=10\)). Conditioned on this history and task-related inputs $\mathbf{c}_t$, it outputs $\mathbf{a}_t\in\mathbb{R}^{29}$ as joint-position residuals relative to the fixed nominal configuration $\mathbf{q}^{\mathrm{nom}}$. The joint-position targets are:
\begin{equation}
\mathbf{q}_t^{\mathrm{des}}
=\mathbf{q}^{\mathrm{nom}}+\mathbf{S}\mathbf{a}_t,
\label{eq:joint_position_target}
\end{equation}
where $\mathbf{S}$ is a diagonal residual-scaling matrix. A joint-level PD controller tracks these targets with torque saturation. The policy is trained using proximal policy optimization (PPO).

\subsubsection{Object-consistent kinematic guidance}
During training, the roll-rate command is integrated into a bounded reference angle to construct the object and grasp references as follows:
\begin{equation}
\begin{aligned}
\phi_t^{\mathrm{ref}}
&=\operatorname{clip}\!\left(
\phi_{t-1}^{\mathrm{ref}}+\Delta t\,\dot{\phi}_t^{\mathrm{cmd}},
\phi_{\min},\phi_{\max}\right),\\
\mathbf{R}_O^{\mathrm{ref}}
&=\mathbf{R}_O^0\mathbf{R}_x(\phi_t^{\mathrm{ref}}),\\
\mathbf{p}_i^{\mathrm{ref}}
&=\mathbf{p}_O^{\mathrm{ref}}+\mathbf{R}_O^{\mathrm{ref}}\mathbf{r}_i(d),
\quad i\in\{L,R\},
\end{aligned}
\label{eq:grasp_reference}
\end{equation}
where $\Delta t$ is the update interval, $[\phi_{\min},\phi_{\max}]$ defines the roll-angle bounds, $\mathbf{R}_O^0$ is the nominal object orientation, and $\mathbf{R}_x(\cdot)$ denotes rotation about the object roll axis. The grasp references are constructed from the reference object position $\mathbf{p}_O^{\mathrm{ref}}$ and grasp offsets $\mathbf{r}_i(d)$, parameterized by grasp spacing $d$. For position regulation, the reference grasp midpoint $\mathbf{p}_G^{\mathrm{ref}}= (\mathbf{p}_L^{\mathrm{ref}}+\mathbf{p}_R^{\mathrm{ref}})/2$ is tracked against the actual grasp midpoint $\mathbf{p}_G$. The target wrist orientations are coupled to the reference object orientation. All kinematic references are expressed in the robot-base frame.

Constrained IK generates feasible grasp configurations while accounting for joint limits, collisions, and posture regularization. The nominal grasp solution defines $\mathbf{q}^{\mathrm{nom}}$, while roll-dependent object and grasp references guide coordinated object tilting through task-space tracking objectives during training. Precomputed IK posture references are retrieved from a lookup table to evaluate the posture reward.

\subsubsection{Load-dependent posture guidance}
To accommodate varying payload masses and object roll angles, we precompute whole-body posture references $\mathbf{q}^{\mathrm{ref}}$ using constrained IK. The optimization accounts for grasp and foot-pose consistency, load-induced CoM shifts, actuator loading, and posture regularization.

During WBC training, the corresponding IK reference is retrieved from a lookup table based on the payload mass and object roll angle and used to compute the posture reward. The observation offset and action baseline remain fixed at $\mathbf{q}^{\mathrm{nom}}$.

\subsubsection{Reference-guided policy learning}
The posture reward regularizes deviations from the load-dependent reference through the weighted error:
\begin{equation}
  E_q=\bigl\|\mathbf{M}\bigl(
  \mathbf{q}-\mathbf{q}^{\mathrm{ref}}
  \bigr)\bigr\|_2^2,
  \label{eq:posture_error}
\end{equation}
where $\mathbf{M}$ is a diagonal joint-weighting matrix, with higher weights on the shoulders and elbows than on the wrists and zeros for unregularized joints. The prior's influence is modulated by $\eta_q=1-0.95|\sin\phi^{\mathrm{ref}}|$, allowing greater posture flexibility during large-angle tilting. Together with object-level tracking and grasp maintenance objectives, this guidance promotes coordinated whole-body motion.

The overall reward formulation incorporates planar velocity, yaw-rate, object orientation, and object roll-rate tracking objectives, augmented by balance rewards and action regularization. Table~\ref{tab:wbc_rewards} summarizes the principal reward terms and their weights. The parameters $s_v$ and $s_\omega$ are squared-error scales in the velocity-tracking rewards. In the object-position reward, $\eta_p$ and $\sigma_p$ are parameterized by the reference roll angle $\phi^{\mathrm{ref}}$.For object orientation tracking, $\phi_{\mathrm{ori}}$ and $\Delta\psi$ denote the object roll angle and wrapped relative yaw error, respectively, evaluated in a gravity-aligned frame that follows the robot heading.

\begin{table}[t]
\centering
\caption{Key reward terms for whole-body policy training}
\label{tab:wbc_rewards}
\setlength{\tabcolsep}{4pt}
\renewcommand{\arraystretch}{1.25}
\resizebox{\columnwidth}{!}{%
\begin{tabular}{@{}lcc@{}}
\toprule
Term & Expression & Weight \\
\midrule

Forward Vel. Tracking
& $\exp\!\left(
-(v_x-v_x^{\mathrm{cmd}})^2/s_v
\right)$
& $4.0$ \\

Lateral Vel. Tracking
& $\exp\!\left(
-(v_y-v_y^{\mathrm{cmd}})^2/s_v
\right)$
& $3.5$ \\

Yaw Rate Tracking
& $\exp\!\left(
-(\omega_z-\omega_z^{\mathrm{cmd}})^2/s_\omega
\right)$
& $6.0$ \\

Object Position Regulation
& $\eta_p\exp\!\left(
-\lVert \mathbf{p}_G-\mathbf{p}_G^{\mathrm{ref}}\rVert_2^2/\sigma_p^2
\right)$
& $4.0$ \\

Object Ori. Tracking
& $\begin{aligned}
\exp\bigl(-&[
\operatorname{wrap}(\phi_{\mathrm{ori}}-\phi^{\mathrm{ref}})^2\\
&+(\Delta\psi)^2]
/\sigma_R^2\bigr)
\end{aligned}$
& $5.0$ \\

Object Roll Rate Tracking
& $\begin{aligned}
\exp\bigl(-&[
(\bar{\omega}_{O/B,x}
-\dot{\phi}^{\mathrm{ref}})^2/\sigma_r^2\\
&+\lVert\bar{\boldsymbol{\omega}}_{O/B,yz}\rVert_2^2
/\sigma_\perp^2]\bigr)
\end{aligned}$
& $4.0$ \\

Object Pitch Penalty & $-\theta_O^2$
& $40.0$ \\

Load-Dependent Posture Prior & $1-\eta_q[1-\exp(-E_q)]$
& $8.0$ \\

\bottomrule
\end{tabular}%
}

\vspace{2pt}
\begin{minipage}{\columnwidth}
\scriptsize \textbf{Note:} $\theta_O$ is the world-frame object pitch. $\bar{\boldsymbol{\omega}}_{O/B}$ is the filtered object angular velocity relative to the base.

\end{minipage}
\end{table}

Training follows a progressive curriculum. It begins with object gravity disabled and gradually introduces varying payload masses, randomized bilateral grasp spacing, and external disturbances. To promote robust adaptation to diverse human partners, we simulate physical human-side support by applying a randomized upward support force at the human grasp point, sampled from a bounded range.

\begingroup
\makeatletter \def\ftype@figure{4} \makeatother
\begin{figure*}[!t]
  \centering
  \includegraphics[width=\textwidth]
  {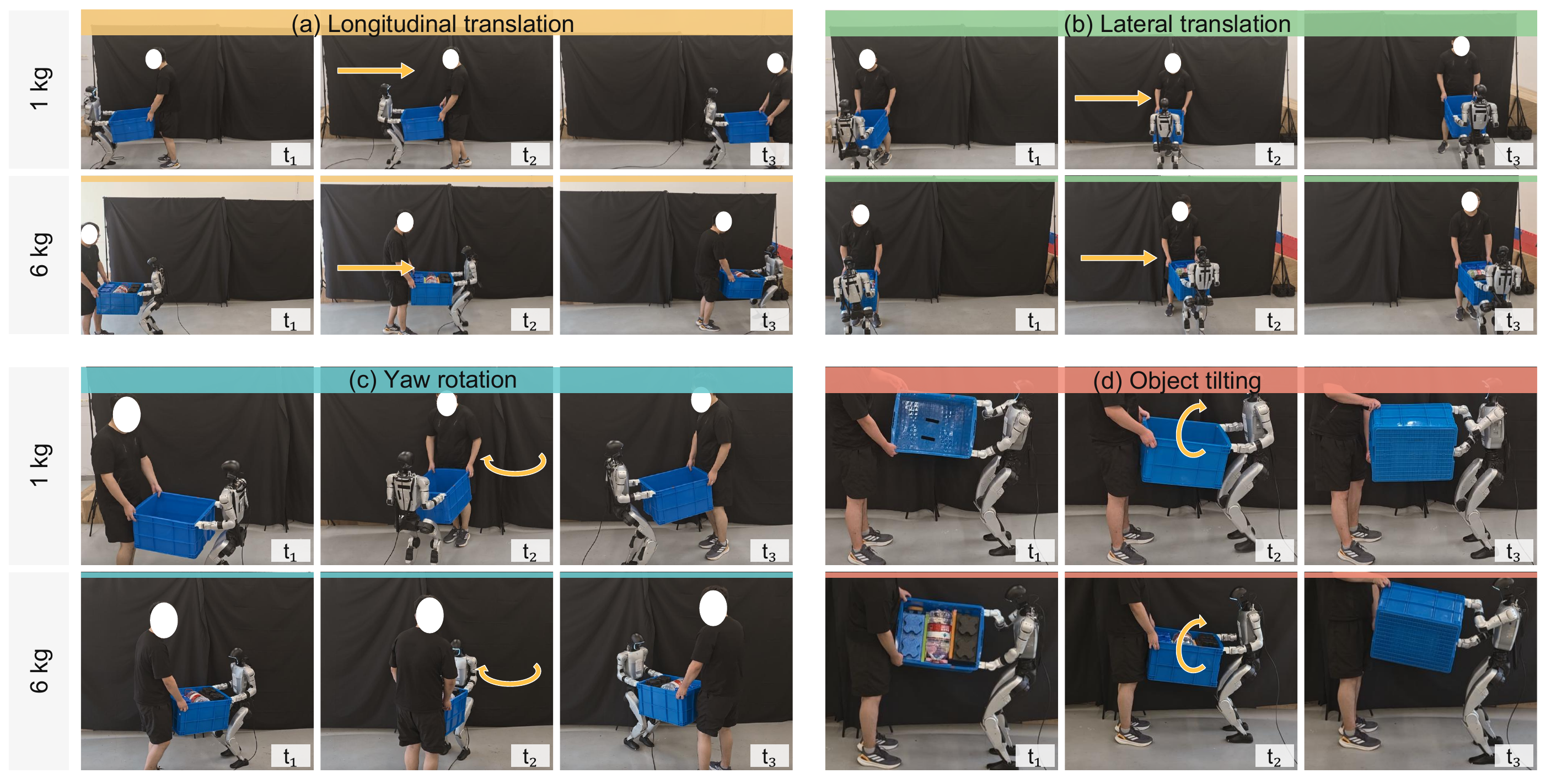}\par
  \nointerlineskip
  \caption{\textbf{Individual motion responses under 1\,kg and 6\,kg load conditions.}\enspace (a) Longitudinal translation. (b) Lateral translation. (c) Yaw rotation. (d) Object roll. The upper and lower rows correspond to the 1\,kg and 6\,kg conditions, respectively. Snapshots at $t_1$, $t_2$, and $t_3$ illustrate the motion progression, with arrows indicating the direction of movement.}
  \label{fig:four-dof-motions}
\end{figure*}
\endgroup

\subsection{Human Motion Intent Learning}
\label{sec:teacher}

We model the motion-guiding component of interaction through a velocity-dependent virtual impedance, allowing guiding forces to decrease as desired and actual motion converge without prescribing fixed force profiles.

\subsubsection{Virtual-impedance interaction modeling}

Human motion intent $\mathbf{c}_t^{\mathrm{des}}$ specifies the desired robot-base planar velocities, robot-base yaw rate, and object roll rate relative to the base, in the same component order, reference points, and coordinate conventions as the WBC command $\mathbf{c}_t$. We sample $\mathbf{c}_t^{\mathrm{sample}}$ over stationary, translational, rotational, and combined motions, then apply a first-order low-pass filter:
\[
\mathbf{c}_t^{\mathrm{des}}=
(1-\alpha)\mathbf{c}_{t-1}^{\mathrm{des}}
+\alpha \mathbf{c}_t^{\mathrm{sample}},
\]
where $\alpha\in(0,1]$.

The virtual-impedance model maps the discrepancy between this target and the filtered realized task-level motion to a bounded interaction wrench:
\begin{equation}
\mathbf{W}_h= \operatorname{sat}\left( \mathbf{K}(\mathbf{c}_t^{\mathrm{des}}-\bar{\mathbf{u}}_t) \right),
\label{eq:human_impedance}
\end{equation}
where $\mathbf{W}_h=[f_{h,x},f_{h,y},\tau_{h,z},\tau_{h,x}]^\top$ consists of planar force components and yaw and roll torques. The subscript $h$ denotes human guidance. The diagonal matrix $\mathbf{K}=\operatorname{diag}(K_v,K_v,K_y,K_r)$ defines the virtual-impedance gains, $\bar{\mathbf{u}}_t$ collects the filtered robot-base planar velocities, robot-base yaw rate, and object roll rate relative to the base, in the same order and coordinate convention as $\mathbf{c}_t^{\mathrm{des}}$. The operator $\operatorname{sat}(\cdot)$ clips each force and torque component to its prescribed lower and upper bounds. Thus, instantaneous interaction wrenches alone do not uniquely determine human motion intent.

For grasp-point evaluation and wrench application at the object CoM, we compute the human-side grasp-point velocity and the equivalent torque as follows:
\begin{equation}
\begin{aligned}
\mathbf{v}_h &= \mathbf{v}_O+\boldsymbol{\omega}_O\times(\mathbf{p}_h-\mathbf{p}_O),\\
\boldsymbol{\tau}_C &= \boldsymbol{\tau}_h+(\mathbf{p}_h-\mathbf{p}_C)\times \mathbf{f}_h,
\end{aligned}
\label{eq:human_grasp_velocity}
\end{equation}
where $\mathbf{v}_O, \boldsymbol{\omega}_O$ are the object's velocities, and $\mathbf{p}_h, \mathbf{p}_O, \mathbf{p}_C$ denote the positions of the human-side grasp point, the object reference point, and the object CoM, respectively. The wrench ($\mathbf{f}_h, \boldsymbol{\tau}_C$) is applied at the object's CoM in simulation.

\subsubsection{Privileged teacher policy}
The teacher observes proprioceptive information, interaction feedback, and object states:
\begin{equation}
\mathbf{o}_t^{\mathrm{T}}\triangleq [\mathbf{o}_t^{\mathrm{prop}},\, \mathbf{w}_t^{\mathrm{grasp}},\, \mathbf{s}_t^{\mathrm{obj}}],
\label{eq:teacher_observation}
\end{equation}
where $\mathbf{w}_t^{\mathrm{grasp}}\in\mathbb{R}^{6}$ is the actual interaction wrench at the robot grasp, and $\mathbf{s}_t^{\mathrm{obj}}$ contains the object position, 6D orientation, and relative linear and angular velocities. A multilayer perceptron (MLP) processes an observation history of length $l_{\mathrm{T}}=10$ (0.20\,s) and generates the command $\mathbf{c}_t^{\mathrm{T}}$ for the frozen WBC. The desired human motion $\mathbf{c}_t^{\mathrm{des}}$ is not included in the teacher observations.

\subsubsection{Intent-supervised policy learning}
The tracking reward evaluates the realized task-level motion in the same coordinates as the human intent:
\begin{equation}
r_t^{\mathrm{track}}= 2\sum_{k\in\mathcal{G}} \exp\left( -\frac{\|\mathbf{e}_{k,t}\|_2^2}{\sigma_k^2} \right),
\label{eq:teacher_tracking}
\end{equation}
where $\mathbf{e}_{k,t}=\bar{\mathbf{u}}_{t}^{(k)} -\mathbf{c}_t^{\mathrm{des},(k)}$, $\mathcal{G}=\{xy,\mathrm{yaw},\mathrm{roll}\}$, and $\sigma_k$ specifies the tracking-error scale for each motion group.

To explicitly supervise motion intent inference, we introduce an auxiliary objective:
\begin{equation}
\begin{aligned}
\mathcal{L}_{\mathrm{intent}} =\operatorname{SmoothL1}\big(
&\tanh(\boldsymbol{\mu}_t),\\
&\operatorname{clip} (\mathbf{D}^{-1}\mathbf{c}_t^{\mathrm{des}},-1,1) \big),
\end{aligned}
\label{eq:teacher_intent}
\end{equation}
where $\boldsymbol{\mu}_t$ is the pre-squashing teacher output and $\mathbf{D}$ is the diagonal matrix of WBC command limits, with $\mathbf{c}_t^{\mathrm{T}}=\mathbf{D}\tanh(\boldsymbol{\mu}_t)$. Unlike the tracking reward, which evaluates realized motion, the auxiliary objective directly supervises the teacher's commands toward the human-intended target.

The teacher policy is then optimized using PPO ($\lambda_{\mathrm{intent}}=5$), with auxiliary terms that penalize command saturation, abrupt command changes, unintended drift, and task failure.

\subsection{Proprioceptive Policy Distillation}
\label{sec:student}

We distill the privileged teacher into a recurrent student using DAgger. The student employs a gated recurrent unit (GRU) to predict motion commands from proprioceptive histories of length $l_{\mathrm{S}}=32$, along with auxiliary estimates of equilibrium and interaction wrenches, $\hat{\mathbf{W}}_{\mathrm{e},t}$ and $\hat{\mathbf{W}}_{\mathrm{i},t}$, respectively. During DAgger rollouts, the teacher provides command labels at student-visited states, while wrench labels are obtained from simulation.

We minimize the following loss:
\begin{equation}
\begin{aligned}
\mathcal{L}_{\mathrm{S}} &= \frac{5}{4} \left\| \tanh(\mathbf{z}_{\mathrm{S}}) - \tanh(\boldsymbol{\mu}) \right\|_2^2 +
\mathcal{L}_{\mathrm{wrench}},\\
\mathcal{L}_{\mathrm{wrench}} &= \operatorname{SmoothL1} \left( \hat{\mathbf{W}}_{\mathrm{e}}, \mathbf{W}_{\mathrm{e}}^{\mathrm{GT}}
\right)\\
&\quad+ \operatorname{SmoothL1} \left( \hat{\mathbf{W}}_{\mathrm{i}}, \mathbf{W}_{\mathrm{i}}^{\mathrm{GT}} \right).
\end{aligned}
\label{eq:student_loss}
\end{equation}
Here, time indices are omitted for brevity. $\mathbf{z}_{\mathrm{S}}$ and $\boldsymbol{\mu}$ denote the pre-squashing outputs of the student and teacher, respectively, and the ground-truth wrenches are normalized simulation measurements. At deployment, only the student and frozen WBC are retained.

\raggedbottom
\section{Experimental Evaluation}
\label{sec:experiments}
We evaluate the framework in simulation and on a Unitree G1 humanoid robot. Simulations assess tracking, task success across payloads, and human guiding effort. Real-world experiments assess whole-body coordination and object tilting during collaborative transport.
\subsection{Experimental Setup}
\label{sec:experimental_setup}

\subsubsection{Training and deployment}

All policies train in Isaac Lab on one NVIDIA RTX 4080 GPU. The WBC uses PPO (10,000 iterations, 4,096 parallel environments). The teacher and student train for 15,000 and 20,000 iterations, respectively. We deploy the distilled student and frozen WBC on a physical humanoid.

\subsubsection{Simulation setup}

In simulation, a virtual human supports half of the object weight to model load sharing between the human and robot, and provides physical guidance through proportional--integral (PI) feedback on grasp-point motion errors $\mathbf{e}_t=\mathbf{g}_{h,t}^{\mathrm{ref}}-\bar{\mathbf{g}}_{h,t}$, yielding an interaction wrench $\mathbf{W}_h^{\mathrm{PI}}= \operatorname{sat}(\mathbf{K}_P\mathbf{e}_t+ \mathbf{K}_I\int_0^t\mathbf{e}_{\tau}\,d\tau)$. Here, $\mathbf{g}_{h,t}^{\mathrm{ref}}$ is the prescribed evaluation reference for human-side grasp-point planar velocities, absolute object yaw rate, and object roll rate relative to the base; $\bar{\mathbf{g}}_{h,t}$ contains the corresponding filtered measurements. These grasp-point references are distinct from the base-referenced intent $\mathbf{c}_t^{\mathrm{des}}$ used in teacher training. The resulting wrench is transformed to the object's CoM and applied as a physical interaction. The integral term allows a nonzero guiding wrench to persist even at small motion errors, enabling sustained physical guidance.

The physics simulation advances at 200\,Hz. Within each comparison, all methods share the same grasp-point references, PI parameters, and disturbance conditions. The guiding wrench is determined by each method's actual motion response, preserving the closed-loop interaction between the virtual human and the robot.

\subsubsection{Real-world setup}

The real-world experiments are conducted with a physical humanoid robot and a human partner guiding the object from the opposite side. The object is rigidly attached to the robot's end effectors. Payloads of 1\,kg and 6\,kg are used for the motion tests. The narrow-doorway scenario uses a typical office doorway, while the container-dumping scenario is conducted in a cluttered environment with obstacles along the transport route.

\begin{table*}[t]
\centering
\caption{Quantitative comparison across payload conditions}
\label{tab:comparison}
\footnotesize
\setlength{\tabcolsep}{3pt}
\renewcommand{\arraystretch}{1.0}
\begin{tabular*}{0.93\textwidth}
{@{\extracolsep{\fill}}llccccc@{}} \toprule Mass & Method & $E_v\downarrow$ (m/s) & $E_{\mathrm{yaw}}\downarrow$ (rad/s) & $E_r\downarrow$ (rad/s)
& $S_p\uparrow$ (\%)
& $S_c\uparrow$ (\%) \\
\midrule

\multirow{6}{*}{2\,kg} & COLA
& \textbf{0.09} & 0.15 & \textcolor{red}{$\times$} & 90 & \textcolor{red}{$\times$} \\
& Torque-based Admittance
& 0.17 & 0.20 & 0.22 & 85 & 75 \\
& Tracking-error Feedback
& 0.20 & 0.23 & 0.27 & 75 & 65 \\
& Direct RL
& 0.14 & 0.16 & 0.18 & 85 & 80 \\
& Ours w/o LB
& 0.11 & 0.12 & \textbf{0.11} & 90 & 85 \\
& \textbf{Ours} & 0.10 & \textbf{0.11} & 0.12
& \textbf{95} & \textbf{90} \\
\midrule

\multirow{6}{*}{4\,kg} & COLA
& 0.13 & \textbf{0.12} & \textcolor{red}{$\times$} & 85 & \textcolor{red}{$\times$} \\
& Torque-based Admittance
& 0.18 & 0.19 & 0.26 & 85 & 70 \\
& Tracking-error Feedback
& 0.19 & 0.27 & 0.29 & 80 & 55 \\
& Direct RL
& 0.16 & 0.18 & 0.17 & 80 & 80 \\
& Ours w/o LB
& 0.14 & 0.16 & 0.20 & 85 & 70 \\
& \textbf{Ours} & \textbf{0.11} & 0.13 & \textbf{0.14}
& \textbf{95} & \textbf{90} \\
\midrule

\multirow{6}{*}{8\,kg} & COLA
& 0.23 & 0.25 & \textcolor{red}{$\times$} & 75 & \textcolor{red}{$\times$} \\
& Torque-based Admittance
& 0.29 & 0.32 & 0.43 & 60 & 50 \\
& Tracking-error Feedback
& 0.27 & 0.39 & 0.45 & 55 & 40 \\
& Direct RL
& 0.21 & 0.28 & 0.31 & 75 & 60 \\
& Ours w/o LB
& 0.26 & 0.31 & 0.40 & 65 & 45 \\
& \textbf{Ours} & \textbf{0.16} & \textbf{0.21} & \textbf{0.24}
& \textbf{85} & \textbf{75} \\
\bottomrule
\end{tabular*}

\par\smallskip
\begin{minipage}{0.93\textwidth}
\footnotesize \textcolor{red}{$\times$} denotes an unsupported task. LB denotes load-dependent posture guidance.
\end{minipage}
\end{table*}

\subsection{Quantitative Analysis}
\label{sec:simulation}

In this section, we quantitatively evaluate the proposed framework through two simulation experiments. The first investigates four-dimensional motion coordination under varying payload conditions, covering planar translation, yaw rotation, object roll, and combined motions. The second examines sustained collaborative transport with a 5\,kg payload, focusing on motion tracking and human guiding effort during motion initiation, sustained transport, and stopping.

\subsubsection{Motion Tracking Across Payloads}

To evaluate motion coordination across varying payloads, we compare our method with four baselines. COLA~\cite{du2026cola} supports planar collaborative transport but does not support commanded object roll. Torque-based Admittance estimates interaction wrenches from model-compensated joint torque feedback, whereas Tracking-error Feedback infers motion commands from joint-state deviations. Direct RL uses our student's recurrent architecture and learns from proprioception without privileged teacher supervision.

To assess the contribution of load-dependent posture guidance, we additionally evaluate Ours w/o LB, whose WBC is trained without this guidance while retaining the other training objectives. Except for COLA, all methods use a four-dimensional command interface and a frozen WBC.

In Table~\ref{tab:comparison}, we report the root mean squared errors (RMSEs) of planar velocity ($E_v$), yaw rate ($E_{\mathrm{yaw}}$), and roll rate ($E_r$), computed by comparing the filtered human-side grasp-point motion with the prescribed grasp-point references $\mathbf{g}_{h,t}^{\mathrm{ref}}$. We also report planar ($S_p$) and combined-motion ($S_c$) success rates over 20 trials per condition. A trial is considered successful if the robot completes the motion without falling
and the maximum absolute tracking error does not exceed 30\% of the reference velocity magnitude.

Across varying payloads, our method achieves competitive tracking accuracy and high task success rates, with a more pronounced advantage under heavier loads. Although COLA achieves slightly lower errors on certain planar metrics at lighter loads, our method achieves the lowest tracking
errors across all three metrics at 8\,kg, with 85\% planar and 75\% combined-motion success
rates. Without load-dependent posture guidance, these success rates drop to 65\% and 45\%,
respectively, accompanied by increased tracking errors. This degradation under heavier loading highlights the role of load-dependent posture guidance in accommodating payload-induced balance challenges during whole-body motion.

\subsubsection{Guiding Effort During Sustained Motion}

To evaluate the effect of intent-aware command generation on sustained guiding force, we compare our method with a PAINT-inspired force-scaled approach~\cite{paint} and a dynamics-based admittance approach. The former trains a student using teacher velocity references scaled by estimated interaction wrench magnitude, whereas the latter maps estimated interaction wrenches to motion commands using explicit dynamics and payload parameters. The PAINT-inspired student shares the same recurrent architecture and proprioceptive inputs as our student.

\begin{figure}[!t]
\centering
\includegraphics[width=\columnwidth,trim=0 23bp 0 0,clip]
{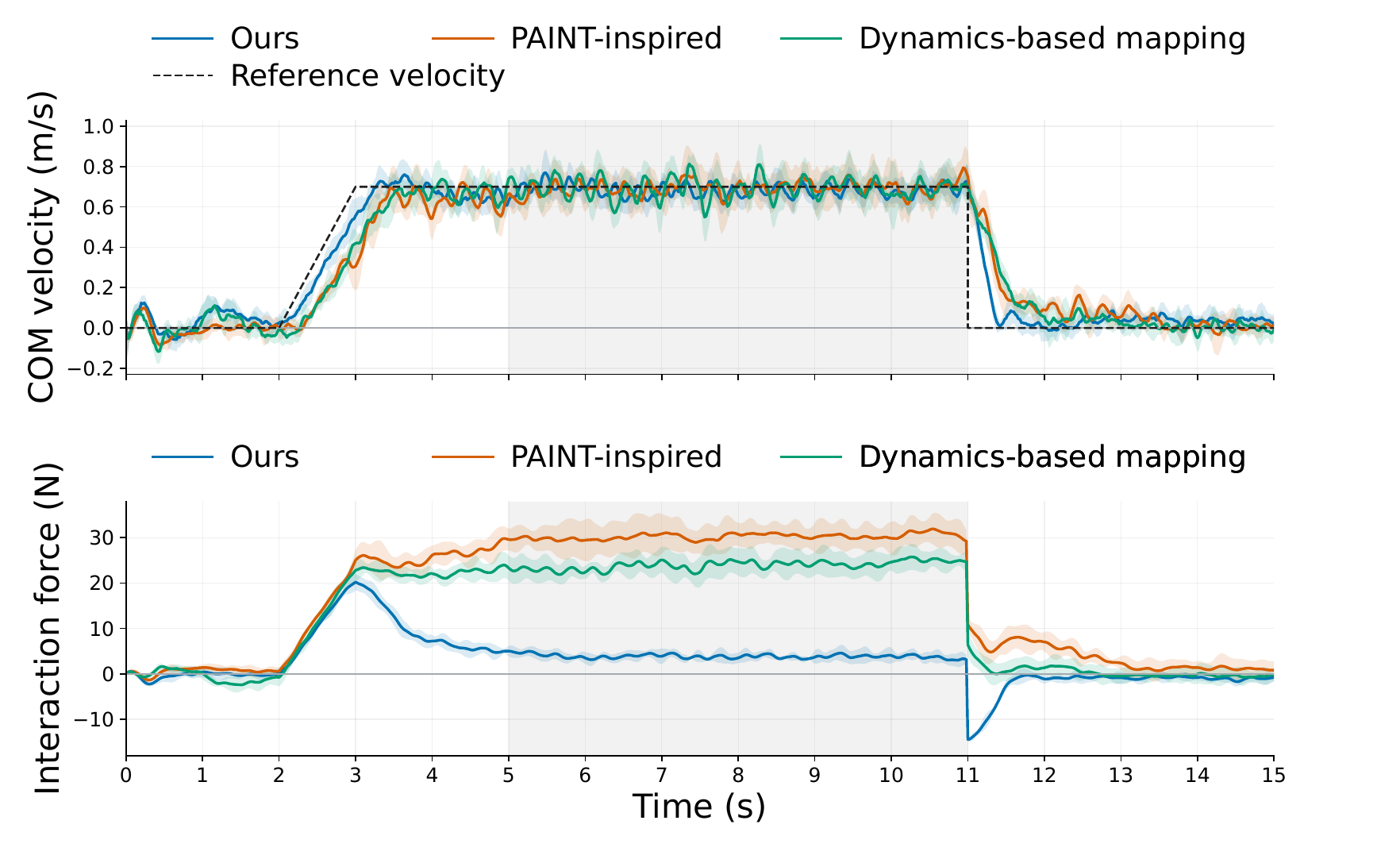}\par
\nointerlineskip
\caption{\textbf{Longitudinal box velocity and human guiding force during motion initiation, sustained transport, and stopping.}\enspace Curves show means over five trials; colored bands indicate $\pm$ one standard deviation. The gray background marks the steady-state evaluation interval (5--11\,s), and the dashed line denotes the reference velocity.}
\label{fig:sustained_guidance}
\end{figure}

As shown in Fig.~\ref{fig:sustained_guidance}, the target longitudinal velocity transitions through four phases: rest (0--2\,s), acceleration to 0.7\,m/s (2--3\,s), sustained transport (3--11\,s), and stopping (11--15\,s). The integral state is retained during stopping to allow residual interaction forces to influence the response.

Table~\ref{tab:sustained_guidance} summarizes the quantitative results over five trials per method. While the virtual human regulates grasp-point motion, the reported velocity-based metrics use the box CoM velocity along the robot-base $x$-axis, denoted by $v_{\mathrm{box}}$. Over $5\,\mathrm{s}\leq t<11\,\mathrm{s}$, we measure the steady-state velocity RMSE $E_{\mathrm{steady}}$ and mean absolute guiding force $F_{\mathrm{steady}}$. The stopping distance is defined as $D_{\mathrm{stop}}=\int_{11}^{15}|v_{\mathrm{box}}(t)|\,dt$. The response time $T_{90}$ is measured from $t=2$\,s to the first time at which $v_{\mathrm{box}}$ remains at or above 0.63\,m/s for at least 0.2\,s.

\begin{table}[!t]
\centering
\caption{Sustained-motion performance over five trials}
\label{tab:sustained_guidance}
\setlength{\tabcolsep}{3pt}
\renewcommand{\arraystretch}{1.15}
\resizebox{\columnwidth}{!}{%
\begin{tabular}{@{}lcccc@{}}
\toprule Method & $E_{\mathrm{steady}}$ & $F_{\mathrm{steady}}$ & $D_{\mathrm{stop}}$
& $T_{90}$ \\
& (m/s) & (N) & (m) & (s) \\
\midrule PAINT-inspired & $0.097\pm0.009$ & $30.27\pm2.97$ & $0.471\pm0.045$
& $1.73\pm0.84$ \\
Dynamics-based & $0.093\pm0.014$ & $23.97\pm2.25$ & $0.446\pm0.030$
& $1.72\pm0.38$ \\
\textbf{Ours} & $\mathbf{0.069\pm0.010}$ & $\mathbf{3.86\pm0.73}$ & $\mathbf{0.331\pm0.070}$
& $\mathbf{1.20\pm0.15}$ \\
\bottomrule
\end{tabular}%
}
\end{table}

Our method exhibits faster motion initiation, reaching the target velocity threshold in 1.20\,s, compared with 1.73\,s and 1.72\,s for the two baselines. More importantly, during sustained transport, it achieves the lowest steady-state velocity error while requiring only 3.86\,N of mean guiding force, compared with 30.27\,N and 23.97\,N for the baselines. These results indicate that the learned policy can maintain the desired motion without relying on large sustained guiding forces. The shorter stopping distance of 0.331\,m further demonstrates its ability to respond to changes in guidance.

\subsection{Real-World Experiments}
\label{sec:real_world}

\subsubsection{Motion-level coordination}

Fig.~\ref{fig:four-dof-motions} illustrates the robot's responses to longitudinal, lateral, yaw, and object-roll guidance under 1\,kg and 6\,kg payloads. While planar motion is primarily achieved through locomotion, object tilting requires coordinated bilateral arm and body adjustments to maintain balance and grasp consistency. Notably, all four motion types are executed using the same learned policies under both payload conditions, without requiring motion-specific controllers. These results show that a unified whole-body controller can coordinate locomotion and object manipulation under physical human guidance.
\begin{figure}[!t]

  \centering
      \includegraphics[width=\columnwidth]
      {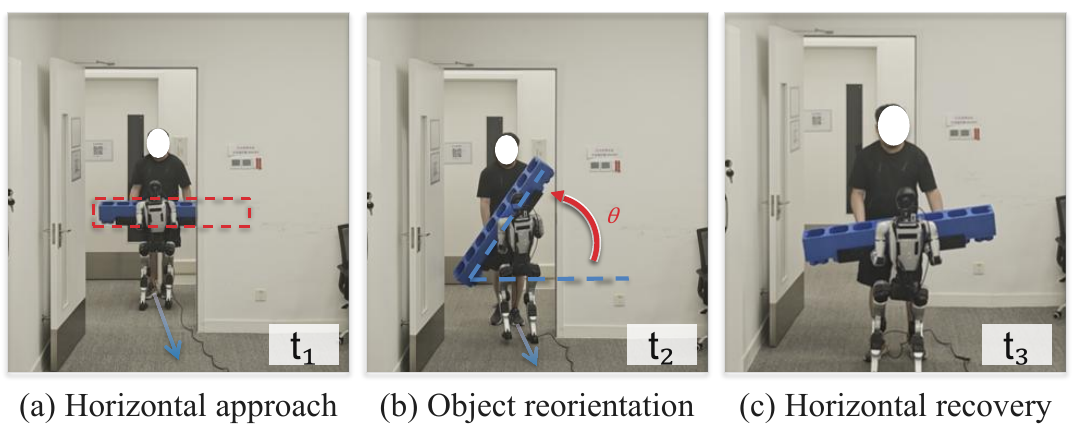}\par
      \nointerlineskip
      \caption{\textbf{Human--humanoid collaborative transport through a narrow doorway.}\enspace (a) Approaching with the object held horizontally. (b) Coordinated turning and object tilting during passage. (c) Restoring the horizontal orientation after clearing the doorway.}
      \label{fig:narrow-passage}
  \par\vspace{\floatsep}
\centering
    \includegraphics[width=\columnwidth,trim=45bp 48bp 151bp 119bp,clip]
    {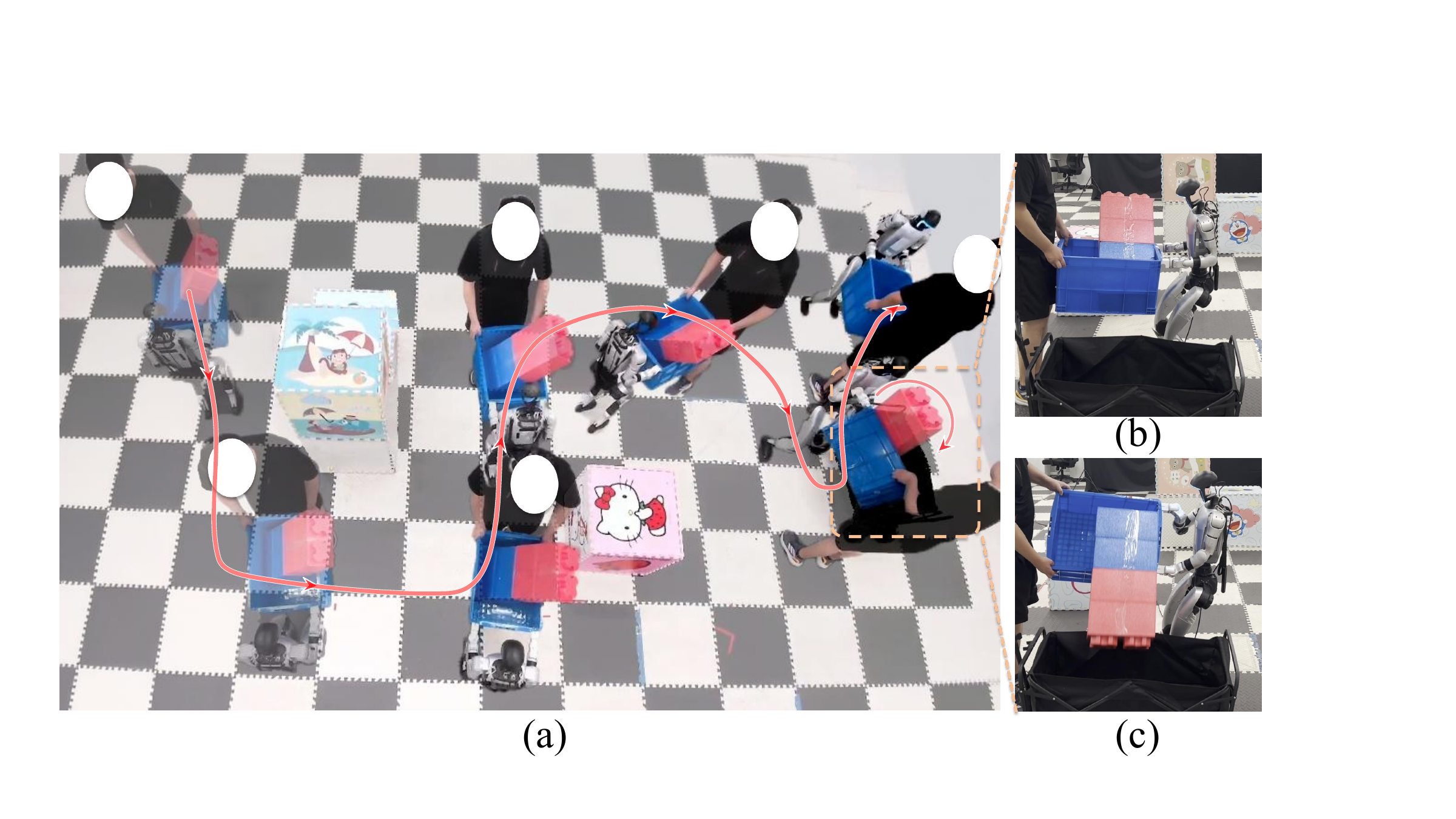}\par
    \nointerlineskip
    \caption{\textbf{Human--humanoid collaborative transport and container dumping.}\enspace (a) Overlaid snapshots of human-guided transport around obstacles, with red arrows indicating the transport sequence. The orange dashed box marks the portion of the transport sequence corresponding to the close-up views in (b) and (c). (b) The container is positioned above the receptacle. (c) At a subsequent stage, the container is tilted to discharge its contents.}
    \label{fig:transport-demo}
\end{figure}

\subsubsection{Task-level collaborative transport}

We further evaluate the integration of locomotion and object manipulation in two real-world collaborative transport tasks. In narrow-doorway traversal (Fig.~\ref{fig:narrow-passage}), the human and robot coordinate translation, turning, and object tilting to maneuver an elongated payload through the constrained passage. This requires object reorientation to be coordinated with locomotion while maintaining balance and grasp consistency. In container dumping (Fig.~\ref{fig:transport-demo}), the robot transports the container around obstacles and positions it above a receptacle before tilting it to discharge its contents. Unlike doorway traversal, object tilting here follows transport and positioning rather than occurring simultaneously with locomotion. Together, these demonstrations highlight the ability of the same learned policies to coordinate locomotion and manipulation in task-dependent ways, suggesting their potential for broader human--humanoid collaborative transport applications.

\section{Conclusion}

The proposed framework enables coordinated object transport and tilting while following human motion intent with substantially reduced sustained guiding force. Object-consistent kinematic guidance facilitates grasp-consistent object tilting, while load-dependent posture guidance supports balanced whole-body motion across varying payloads. At 8\,kg, the latter improves the combined-motion success rate from 45\% to 75\%. Intent-aware command generation sustains the desired motion as guiding forces diminish, reducing steady-state guiding force to 3.86\,N versus 23.97--30.27\,N for the baselines. Future work will extend the framework to multi-humanoid transport with coordinated whole-body motion and shared object manipulation.

\FloatBarrier
\bibliographystyle{IEEEtran}
\bibliography{references}
\end{document}